\documentclass[11pt]{article}
\usepackage{paclic32}
\usepackage{times}
\usepackage{latexsym}
\usepackage{amssymb}
\usepackage{amsmath}
\usepackage{amsthm}
\usepackage{multirow}
\usepackage{url}
\usepackage{graphicx}
\usepackage{color}

\usepackage{footnote}
\makesavenoteenv{tabular}
\makesavenoteenv{table*}

\let\oldhat\hat
\renewcommand{\vec}[1]{\mathbf{#1}}
\renewcommand{\hat}[1]{\oldhat{\mathbf{#1}}}

\title{Automatic Identification of Indicators of Compromise \\ 
using Neural-Based Sequence Labelling}

\author{\begin{tabular}{cccc}
  Shengping Zhou & Zi Long & Lianzhi Tan & Hao Guo
\end{tabular} \\
  Mobile Internet Group, Tencent \\
  Han’s Laser Technology Centre , Shennan Ave No.9988, \\
  Nanshan District, Shenzhen City, Guangdong Province, 518057, China
} 
\date{}

\begin{document}
\maketitle

\begin{abstract}
  Indicators of Compromise (IOCs) are artifacts observed on a network or in an operating system that can be utilized to indicate a computer intrusion and detect cyber-attacks in an early stage. Thus, they exert an important role in the field of cybersecurity. However, state-of-the-art IOCs detection systems rely heavily on hand-crafted features with expert knowledge of cybersecurity, and require a large amount of supervised training corpora to train an IOC classifier. In this paper, we propose using a neural-based sequence labelling model to identify IOCs automatically from reports on cybersecurity without expert knowledge of cybersecurity. Our work is the first to apply an end-to-end sequence labelling to the task in IOCs identification. By using an attention mechanism and several token spelling features, we find that the proposed model is capable of identifying the low frequency IOCs from long sentences contained in cybersecurity reports. Experiments show that the proposed model outperforms other sequence labelling models, achieving over 88\% average F1-score.
\end{abstract}

\section{Introduction}
\label{sec:intro}

 Indicators of Compromise (IOCs) are forensic artifacts that are used as signs when a system has been compromised by an attacker or infected with a particular piece of malware. To be specific, IOCs are composed of some combinations of virus signatures, IPs, URLs or domain names of botnets, MD5 hashes of attack files, etc. They are frequently described in cybersecurity reports, much of which are written in unstructured text, describing attack tactics, technique and procedures, and can be utilized for early detection of future attack attempts by using intrusion detection systems and antivirus software. 
 With the rapid evolvement of cyber threats, the IOC data are produced at a high volume and velocity every day, which makes it increasingly hard for human to gather and manage them. 
 A number of systems are proposed to help discover and gather malicious information and IOCs from various types of data sources~\cite{ZhuZ16,LiaoX16,Husari17,HuangC17,Kwon17,ZhuZ18}. 
 However, most of them identify IOCs by using human-crafted features that heavily rely on specific language knowledge such as dependency structure, and they often have to be pre-defined by experts in the field of the cybersecurity. Furthermore, they need a large amount of annotated data used as the training data to train an IOC classifier. Those training data are frequently difficult to be crowed-sourced, because non-experts can hardly distinguish IOCs from those non-malicious IPs or URLs. Thus, it is a time-consuming and laborious task to construct such a system.
 
 In this paper, we propose using an end-to-end neural sequence labelling model to fully automate the process of IOCs identification. Among the previous studies of the neural sequence labelling task, Huang et al.~.\shortcite{HuangZ15} proposed using a sequence labelling model based on the bidirectional long short-term memory (LSTM)~\cite{Hochreiter97} for the task of name entity recognition (NER). Chiu and Nichols~\shortcite{Chiu16} and Lample et al.~.\shortcite{Lample16} proposed integrating LSTM encoders with character embedding and the neural sequence labelling model to achieve a state-of-the-art performance on the task of NER. Besides, Dernoncourt et al.~\shortcite{Dernoncourt17} and Jiang et al.~\shortcite{JiangZ17} proposed applying the neural sequence labelling model to the task of de-identification of medical records. To the best of our knowledge, we are the first to apply an end-to-end sequence labelling to the task of IOCs identification in cybersecurity.
 
 The proposed approach is on the basis of an artificial neural networks (ANN) with bidirectional LSTMs and Conditional Random Fields (CRF)~\cite{Lafferty01}, which shows promising results for named entity recognition~\cite{Lample16,Dernoncourt17}. 
 Considering that sentences of cybersecurity reports are different from those news articles and patient notes, which always contain a large number of tokens, and sometimes lists of IOCs with little context, we make use of an attention mechanism that helps LSTM to encode the input sequence accurately. We further introduce several token spelling features to the ANN model so that the proposed model can perform well even with a very small amount of training corpora. Based on the results of our experiments on English cybersecurity reports, our proposed approach achieved an average precision of 90.4\% and the recall of 87.2\%.

\begin{figure*}[tb]
\centering
\includegraphics[scale=0.38]{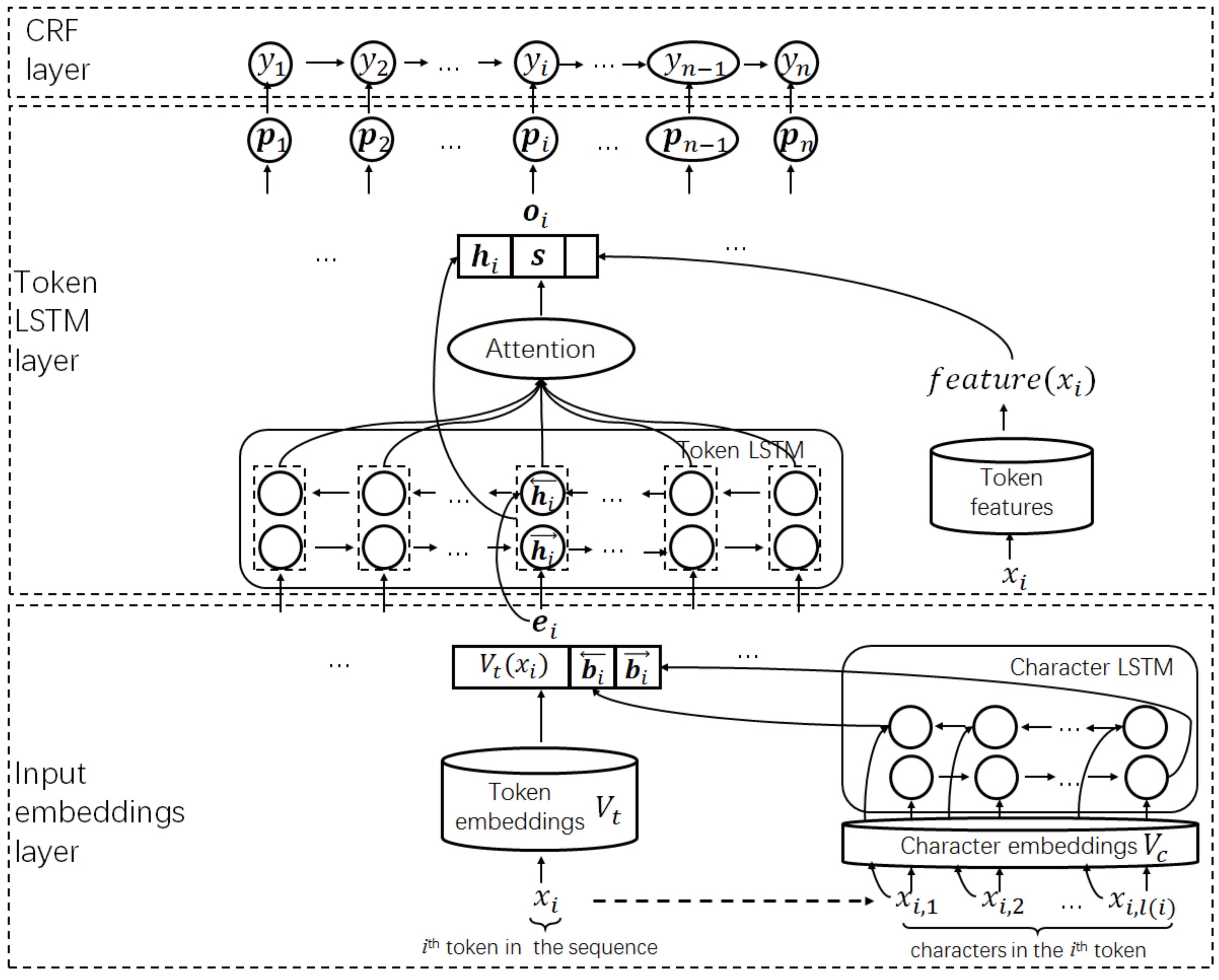}
\caption{ANN model of sequence labeling for IOCs automatic identification}
\label{fig:model}
\end{figure*}

\section{Model}
\label{sec:model}

 Figure~\ref{fig:model} shows the components (layers) of the proposed neural network architecture.

\subsection{Input Embedding Layer}
\label{sec:embedding}

 The input embedding layers takes a token as input and outputs its vector representation. Similar to the work of Lample et al.~\shortcite{Lample16}, the output vector results from the concatenation of two different types of embeddin: the first one directly maps a token to a vector, while the second one outputs a character-level token encoder.

 
 As shown in Figure~\ref{fig:model}, given an input sequence of tokens $x_{1}, \ldots, x_{n}$, each token $x_{i}$ ($i = 1, \ldots, n$) is mapped to a token embedding $V_t(x_i)$ with the mapping of token embedding $V_t(\cdot)$. The token embedding is pre-trained on large unlabeled datasets, and is learned jointly with the rest of the model. Then, let $x_{i,1}, \ldots, x_{i,l(i)}$ be the sequence of characters that comprise the token $x_{i}$, where $l(i)$ is the number of characters in $x_{i}$. Each character $x_{i,j}$ ($j=1, \ldots, l(i)$) is mapped to a character embedding $V_c(x_{i,j})$ using the mapping of character embedding $V_c(\cdot)$. 
 The character embedding is randomly initialized and also jointly learned during the training process.
 Then the vector $V_c(x_{i,j})$ is passed to a bidirectional LSTM, which outputs a forward character-based token embedding $\overrightarrow{b_i}$ and a backward embedding $\overleftarrow{b_i}$. 
 Finally, the output $\vec{e}_{i}$ of the input embedding layer for the $i^{th}$ token $x_{i}$ is the concatenation of the token embedding $V_t(x_i)$ and the character-based token embeddings $\overrightarrow{\vec{b}_i}$, $\overleftarrow{\vec{b}_i}$.
 %
 
\subsection{Token LSTM Layer}
\label{sec:lstm}

 The token LSTM layer takes the sequence of embeddings $\vec{e}_{i}$ ($i =1,\ldots,n$) as input, and outputs a sequence $\vec{p}_{i}(i =1,\ldots,n)$, where the $t^{th}$ element of $\vec{p}_i$ represents the probability that the $i^{th}$ token has the label $t$.
 
 Different from the previous work of name entity recognition in news articles or patient notes, sentences from a cybersecurity report often contain a large number of tokens as well as lists of IOCs with little context, making it much more difficult for LSTM to encode the input sentence correctly. Considering that tokens cannot contribute equally to the representation of the input sequence, we introduce an attention mechanism to extract such tokens that are crucial to the meaning of the sentence. Then, we aggregate the representation of those informative words to form the vector of the input sequence. The attention mechanism is similar to the one proposed by Yang et al.~\shortcite{Yang16}, which is defined as follows:
 \begin{eqnarray}
  \vec{u}_{i} & = & \tanh(W_{w}\vec{h}_{i} + \vec{b}_{w}) \nonumber \\
  \alpha_{i} & = & \frac{\exp(\vec{u}_{i}^\top\vec{u}_{w})}{\sum_{i}\exp(\vec{u}_{i}^\top\vec{u}_{w})} \nonumber \\
  \vec{s} & = & \sum_{i} \alpha_{i}\vec{h}_{i} \nonumber
 \end{eqnarray}
 That is to say, we first compute the $\vec{u}_i$ as a hidden representation of the hidden states of LSTM $\vec{h}_{i}$ for $i^{th}$ input token, i.e., $\vec{h}_{i} = [\overrightarrow{\vec{h}}_i; \overleftarrow{\vec{h}}_i]$.
 Then, we measure the importance of the $i^{th}$ token with a trainable vector $\vec{u}_w$
 and get a normalized importance weight $\alpha_{i}$ through a softmax function. After that, the sentence vector $\vec{s}$ is computed as a weight sum of $\vec{h}_{i}$ ($i=1,\ldots,n$). Here, weight matrix $W_{w}$, bias $\vec{b}_{w}$ and vector $\vec{u}_w$ are randomly initialized and jointly learned during the training process. Noted that each input sentence merely has one sentence vector $\vec{s}$ as its weighted representation, and $\vec{s}$ is then used as a part of each output $\vec{o}_{i}$ ($i=1,\ldots,n$).
 
 Furthermore, we introduce some spelling features to defined IOCs to improve the performance of the proposed model on a very small amount of training data. Here we define several token spelling features and map each token $x_i$ ($x=1, \ldots, n$) to a vector $feature(x_i)$, where the $q^{th}$ element of $feature(x_i)$ represents the value of the $q^{th}$ feature of token $x_i$. 
 Noted that 
 the hand-crafted token spelling features are only applied to initialization, and values of features are jointly learning during the process of training.
 
 As shown in Figure~\ref{fig:model}, the vector $\vec{o}_{i}$ ($i=1,\ldots,n$) is a concatenation of the $i^{th}$ LSTM hidden states $\vec{h}_{i}$, the sentence vector $\vec{v}$ and the feature vector $feature(x_i)$. Each vector $\vec{o}_{i}$ is then given to a feed-forward neural network with one hidden layer, which outputs the corresponding probability vector $\vec{p}_{i}$.

\subsection{CRF Layer}
\label{sec:predict}

\begin{table*}
 \centering
 \caption{Statistics of datasets}
 \label{tab:dataset}
 \begin{tabular}{|p{4cm}||c|c|c|}
    \hline
     & training & validation set & test set \\ \hline\hline
    attacker & 5,304 & 1,067 & 1,609 \\ \hline
    attack method & 2,737 & 610 & 882 \\ \hline
    attack target & 3,055 & 1,055 & 695 \\ \hline
    domain & 6,443 & 1,054 & 1,701 \\ \hline
    e-mail address & 1,284 & 154 & 222 \\ \hline
    file hash & 10,367 & 2,055 & 2,459 \\ \hline
    file information & 4,353 & 1,024 & 1,131 \\ \hline
    IPv4 & 3,012 & 729 & 819 \\ \hline
    malware & 7,317 & 1,585 & 1,974 \\ \hline
    URL & 1,849 & 105 & 156 \\ \hline
    vulnerability & 1,557 & 309 & 359 \\ \hline \hline
    tokens & 1,169,896 & 253,336 & 350,406 \\ \hline
    paragraphs & 6,702 & 1,453 & 2,110 \\ \hline
    articles & 250 & 70 & 70 \\ \hline
 \end{tabular}
\end{table*}

 We also introduce a CRF layer to output the most likely sequence of predicted labels. 
 The score of a label sequence $y_{i}(i =1,\ldots,n)$ is defined as the sum of the probabilities of unigram labels and the bigram label transition probabilities:
 \begin{eqnarray}
     s(y) = \sum^n_{i=1} p_i[y_i] + \sum^n_{i=2} T[y_{y-1}, y_{y}] \nonumber
 \end{eqnarray}
 where $T$ is a matrix that contains the transition probabilities of two subsequent labels. Vectors $\vec{p}_{i}$ is the output of the token LSTM layer, and $T[g,h]$ is the probability that a token with label $g$ is followed by a token with the label $h$. Subsequently, these scores are turned into probabilities of the label sequence by taking a softmax function over all possible label sequences. 

\section{Evaluation}
\label{sec:eva}

\subsection{Datasets}
\label{sec:data}

 As English data, we crawled 687 cybersecurity articles from a collection of advanced persistent threats (APT) reports which are published from 2008 to 2018\footnote{
    \url{https://github.com/CyberMonitor/APT_Cy} \url{berCriminal_Campagin_Collections}
 }. All of those cybersecurity articles are used to train the word embedding. 
 Afterwards, we randomly selected 370 articles, and manually annotate the IOCs contained in the articles. Among the selected articles, we randomly select 70 articles as the validation set and 70 articles as the test set; the remaining articles are used for the training set. Table~\ref{tab:dataset} shows statistics of the datasets. The output labels are annotated with the BIO (which stands for ``Begin'', ``Inside'' and ``Outside'') scheme.

\subsection{Token Spelling Features}
\label{sec:feature}

 Table~\ref{tab:feature} lists all the spelling features for a given token.

 Values of features are then formed as a vector, and are concatenated with the LSTM hidden state vector and the sentence vector of attention in the token LSTM layer\footnote{
   We concatenate the feature vector at different locations in the proposed model, i.e., the input of the token LSTM layer ($e_{i} = [V_t(x_i);\overrightarrow{\vec{b}_i};\overleftarrow{\vec{b}_i};feature(x_i)]$), the hidden state of the token LSTM ($\vec{h}_{i}; = [\overrightarrow{\vec{h}}_i;\overleftarrow{\vec{h}}_i;feature(x_i)]$), and the output of token LSTM ($\vec{o}_i=[\vec{h}_{i};\vec{s};feature(x_i)]$). Among them the third alternative achieved the best performance.
   We speculate that spelling features played an important role in the task of IOCs identification and feature vectors near the output layer was able to improve the performance more significantly than those at other locations.
 } as shown in Section~\ref{sec:lstm}.

\begin{table*}[ht]
\centering
\caption{token spelling features}
\label{tab:feature}
\begin{tabular}{|p{4cm}||p{11cm}|}
    \hline
    features & definition \\ \hline \hline
    IPv4 feature & Return 1 when the token contains 4 digits ($<$256) and maybe a digit as the port. \\ \hline
    domain feature & Return 1 when the token has an identified top-level domain\footnote{
       \url{http://data.iana.org/TLD/tlds-alpha-by} \url{-domain.txt}
      }. \\ \hline
    hash feature & Return 1 when the token is a hexadecimal string with the length of 32, 40 or 64. \\ \hline
    URL feature & Return 1 when the token matches a regular expression \\ 
     & $\verb|http(s)?:\\[0-9a-zA-Z_\.\-\\]+|$. \\ \hline
    vulnerability feature & Return 1 when the token matches a regular expression \\ 
     & $\verb|CVE-[0-9]{4}-[0-9]{4,6}|$. \\ \hline
    file information feature & Return 1 when the token matches a regular expression \\
     & $\verb|[a-zA-Z]{1}:\\[0-9a-zA-Z_\.\-\\]+|$ \\ \hline
    e-mail address feature & Return 1 when the token contains a string that matches a regular expression $\verb|[0-9a-zA-Z_\.\-]+|$, the ``@'' and a domain. \\ \hline
    malware feature & Return 1 when the token starts with the malware type, and contains the common delimiter\footnote{
       \url{https://www.microsoft.com/en-us/securit} \url{y/portal/mmpc/shared/malwarenaming.aspx}
      }. \\ \hline \hline
    other features & Return 1 when the token contains digits. \\ \cline{2-2}
     & Return 1 when the token merely consists of digits. \\ \cline{2-2}
     & Return 1 when the token contains alphabets. \\ \cline{2-2}
     & Return 1 when the token merely consists of alphabets. \\ \cline{2-2}
     & Return 1 when the token contains both digits and alphabets. \\ \cline{2-2}
     & Return 1 when the token merely consists of digits and alphabets. \\ \cline{2-2}
     & Return 1 when the token contains ``.'' and return the number of ``.'' contained. \\ \cline{2-2}
     & Return 1 when the token contains ``$\verb|\|$'' and return the number of ``$\verb|\|$'' contained. \\ \cline{2-2}
     & Return 1 when the token contains ``@'' and return the number of ``@'' contained. \\ \cline{2-2}
     & Return 1 when the token contains ``:'' and return the number of ``:'' contained. \\ \hline
 \end{tabular}
\end{table*}

\begin{table*}[ht]
 \centering
 \caption{evaluation results (micro average for 11 labels) }
 \label{tab:result}
 \begin{tabular}{|p{5cm}||c|c|c|}
    \hline
    Models & Precision & Recall & F1-score \\ \hline \hline
    Baseline & 47.1 & 58.8 & 52.3 \\ \hline\hline
    Huang et al.~\shortcite{HuangZ15} & 64.8 & 33.6 & 51.6 \\ \hline
    Lample et al.~\shortcite{Lample16} & 83.0 & 75.2 & 78.9 \\ \hline
    Rei et al.~\shortcite{ReiM16} & 81.6 & 74.5 & 77.9 \\ \hline \hline
    Our model & {\bf 90.4} & {\bf 87.2} & {\bf 88.8}  \\ \hline
 \end{tabular}
\end{table*}

\begin{table*}[ht]
 \centering
 \caption{evaluation results for each labels (Precision / Recall / F1-score) }
 \label{tab:label_result}
 \begin{tabular}{|p{3cm}||c||c|c|}
    \hline
     & Lample et al.~\shortcite{Lample16}  &  Our model & Our model without\\ 
     &  &    &  additional features \\ \hline\hline
    attacker & 89.2 / 66.5 / 78.1 & {\bf 94.7} / {\bf 73.6} / {\bf 82.8} & 94.2 / 70.6 / 80.7 \\ \hline
    attack method & 78.0 / 67.7 / 74.8 & 72.5 / {\bf 92.0} / {\bf 91.1} & {\bf 93.2} / 85.8 / 89.3  \\ \hline
    attack target & 81.2 / 66.5 / 76.1 & {\bf 90.2} / {\bf 87.8} / {\bf 89.0} & 88.6 / 86.4 / 87.5 \\ \hline
    domain & 67.0 / 64.3 / 65.6 & {\bf 91.2} / {\bf 95.3} / {\bf 93.2} & 82.9 / 61.3 / 70.5 \\ \hline
    e-mail address & 63.8 / 20.7 / 31.3 & {\bf 93.8} / {\bf 92.5} / {\bf 93.2} & 83.3 / 40.3 / 54.3 \\ \hline
    file hash & 85.9 / 97.7 / 91.4 & 88.3 / {\bf 99.9} / {\bf 93.7} & {\bf 89.0} / 98.5 / 93.5 \\ \hline
    file information & 72.4 / 67.7 / 70.0 & 78.7 / {\bf 80.2} / {\bf 79.4} & {\bf 83.3} / 58.9 / 69.0 \\ \hline
    IPv4 & 77.6 / 76.1 / 76.9 & 83.7 / {\bf 96.3} / {\bf 89.6} & 83.6 / 95.2 / 89.0 \\ \hline
    malware & 74.4 / 61.4 / 67.2 & {\bf 95.6} / {\bf 56.9} / {\bf 71.3} & 85.6 / 56.0 / 67.7 \\ \hline
    URL & 98.2 / 94.1 / 96.1 & {\bf 99.2} / {\bf 94.1} / {\bf 96.6} & 98.2 / 93.1 / 95.6 \\ \hline
    vulnerability & 87.7 / 88.9 / 88.3 & {\bf 95.5}  / {\bf 95.5} / {\bf 95.5} & 95.3 / 94.1 / 94.7 \\ \hline \hline
    micro average & 83.0 / 75.2 / 78.9 & {\bf 90.4} / {\bf 87.2} / {\bf 88.8} & 90.0 / 78.8 / 84.0 \\ \hline
 \end{tabular}
\end{table*}

\subsection{Training Details}
\label{sec:train}

 For pre-trained token embedding, we apply word2vec~\cite{Mikolov13} to all crawled 687 English APT reports described in Section~\ref{sec:data} using a window size of 8, a minimum vocabulary count of 1, and 15 iterations. The negative sampling number of word2vec is set to 8 and the model type is skip-gram. The dimension of the output token embedding is set to 100.
 
 The ANN model is trained with the stochastic gradient descent to update all parameters, i.e., token embedding, character embedding, parameters of bidirectional LSTMs, weights of attention, token features, and transition probabilities of CRF layers at each gradient step. 
 For regularization, the dropout is applied to the character-enhanced token embedding before the token LSTM layer. 
 Further training details are given below: 
 (a) Dimensions of character embedding, hidden states of character-based token embedding LSTM, and hidden states of label prediction LSTM are set to 25, 25, and 100, respectively. 
 (b) All of the LSTM’s parameters are initialized with a uniform distribution ranging from -1 to 1.
 (c) We train our model with a fixed learning rate of 0.005. We compute the average F1-score of the validation set by the use of the currently produced model after every epoch had been trained, and stop the training process when the average F1-score of validation set fails to increase during the last ten epochs. We train our model for, if we do not early stop the training process, 100 epochs as the maximum number. 
 (d) We rescale the normalized gradient to ensure that its norm does not exceed 5. 
 (e) The dropout probability is set to 5. 
 
 We train the ANN model on the training set. The training time is around 10 hours when using the described parameters on an 8-CPU machine.

\subsection{Results}
\label{sec:result}

 As shown in Table~\ref{tab:result}, we report the micro average of precision, recall and F1-score for all 11 types of labels for a baseline as well as the proposed model.
 As the baseline, we simply judge the input token as IOCs on the basis of the spelling features described in Section~\ref{sec:feature}\footnote{
  For types of ``attacker'', ``attack method'' and ``attack target'', only tokens that appeared in the training set are identified by the baseline.
 } . As presented in Table~\ref{tab:result}, the score obtained by the proposed model is clearly higher than the baseline.

\begin{figure}[ht]
\caption{Impact of the training set size on F1-score}
\label{fig:data_size}
\centering
\includegraphics[scale=0.53]{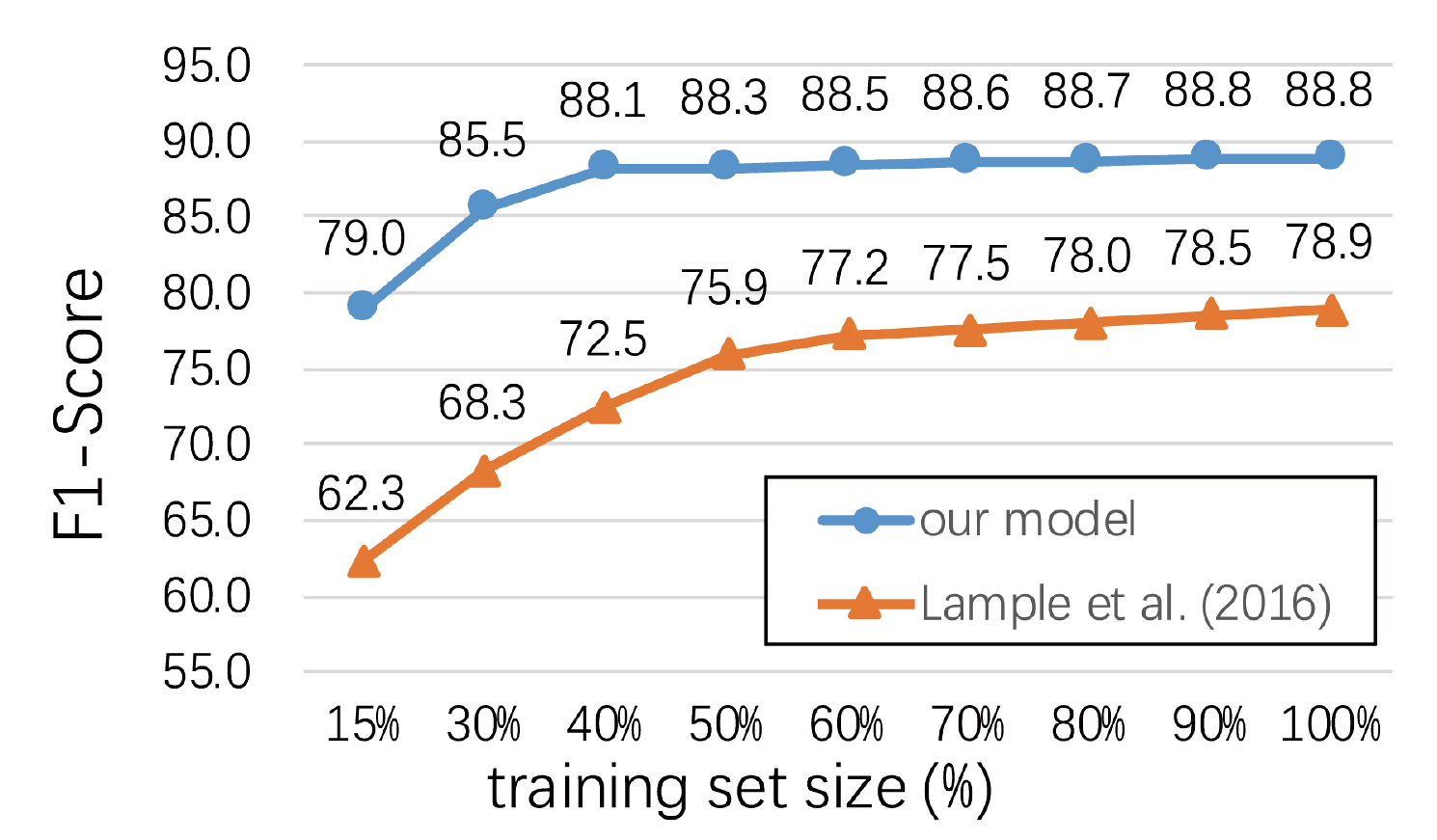}
\end{figure}
\begin{table*}[tb]
\caption{Examples of correct identification by the proposed model }
\label{tab:correct_example}
\centering
\includegraphics[scale=0.42]{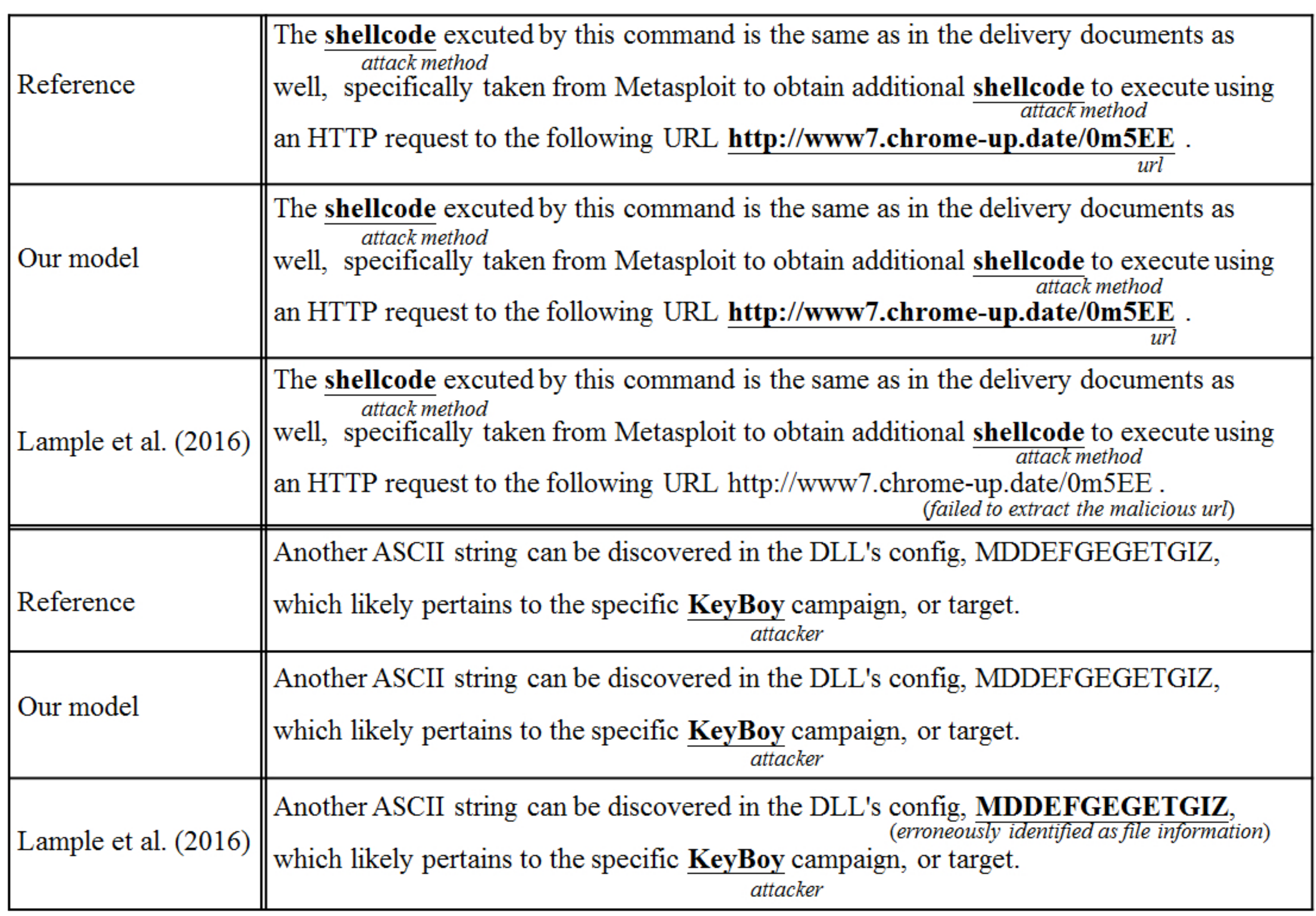}
\end{table*}

 Furthermore, we quantitatively compare our study with other works of name entity recognition, i.e., the work of Huang et al.~\shortcite{HuangZ15}, the work of Lample et al.~\shortcite{Lample16} and the work of Rei et al.~\shortcite{ReiM16}.
 We train these models by employing the same training set and training parameters as the proposed model. As shown in Table~\ref{tab:result}, the proposed model obtains the highest precision, recall and F1-score than other NER models in the task of IOCs extraction. Compared with the second-best model of Lample et al.~\shortcite{Lample16}, the performance gain of the proposed model is approximately 7.4\% of precision, 12.0\% of recall and 9.9\% of the F1-score. Table~\ref{tab:label_result} shows the comparison of scores of each label between Lample et al.~\shortcite{Lample16} and out proposed models. Based on Table~\ref{tab:label_result}, the proposed model achieves better performance for every label, which proves the effectiveness of the proposed model.

 To prove the effectiveness of the spelling features, we further compare the proposed model with the model without spelling features. As shown in Table~\ref{tab:label_result}, the model with features obtains slightly lower scores of precision for some types of labels, and obviously higher scores of recall and F1-score for all types of labels. 
 This is mainly because parts of the IOCs in the test set are newly introduced and appear infrequently. Therefore, the model without spelling features fails to identify those low frequency IOCs for the lack of context information, while the proposed model correctly identifies those IOCs using spelling features as extra information. 
 However, model with hand-crafted spelling features may cause more extraction of false positives, i.e., tokens that have similar to IOCs but are not malicious. The problem is expected to be solved by the introduction of some context features for IOCs tend to be described in a simple and straightforward manner with a fixed set of context tokens~\cite{LiaoX16}.
 
 Moreover, Figure~\ref{fig:data_size} shows the impact of the training set size on the performance of the models. When the training set size is rather limited, the proposed model achieves a greater improvement on F1-score the Lample et al.~\shortcite{Lample16}, since the proposed model uses spelling features as extra information to identify IOCs that have little context information.

 Table~\ref{tab:correct_example} compares several examples of correct IOC extraction produced by the proposed model with one by the work of Lample et al.~\shortcite{Lample16}. In the first example, the model of Lample et al.~\shortcite{Lample16} fails to identify the malicious URL ``http://www7.chrome-up.date/0m5EE'', because the token only appears in the test set and consists of several parts that are uncommon for URLs, such as ``www7'' and ``date'', and thus both the token embedding and the character embedding lack proper information to represent the token as a malicious URL. The proposed model correctly identifies the URL, where the token is defined as a URL by spelling features and is then identified as a malicious URL by the use of the context information. 
 In the second example, token ``MDDEFGEGETGIZ'' is erroneously identified as the name of a malicious file by the model of Lample et al.~\shortcite{Lample16} because of the context ``DLL's config'' before the token that tends to co-occur with names of files. The token is correctly identified by the proposed model, because the token fails to match the regular expression of file information, and is consequently not considered as a name of a malicious file.

\section{Related Work}
\label{sec:related}

\paragraph{NLP in cybersecurity}
Few references in cyber security utilize natural language processing.  Neuhaus and Zimmermann~\shortcite{Neuhaus10} analyze the trend of vulnerability by applying latent Dirichlet allocation to vulnerability description. Liao et al.~\shortcite{LiaoX16} put forward a system to automatically extract IOC items from blog posts. Husari et al.~\shortcite{Husari17} proposed a system that automatically extracted threat actions from unstructured threat intelligence reports by utilizing a pre-defined ontology. 
A concurrent work by Zhu et al.~\shortcite{ZhuZ18} automatically extracted IOC data from security technical articles and further categorized them into different stages of malicious campaigns.
All of those systems consist of several components that rely heavily on manually defined rules, while our proposed model is an end-to-end model using word embedding and spelling features as input, which is more general and applicable to a broader area. 

\paragraph{Neural NER models}
There are amount of ANN-based works 
in the area of named entity recognition. Collobert et al.~\shortcite{Collobert11} described one of the first task-independent neural tagging models on the basis of convolutional neural networks. Hammerton~\shortcite{Hammerton2003} first proposed NER with LSTM.  
Huang et al.~\shortcite{HuangZ15} proposed a bidirectional LSTM model with a CRF layer, including hand-crafted features specialized for the task of NER. Lample et al.~\shortcite{Lample16} described a model where the character-level representation was concatenated with word embedding and Rei et al.~\shortcite{ReiM16} improved the model by introducing an attention mechanism to the character-level representations. 
Dernoncourt et al.~\shortcite{Dernoncourt17} proposed applying the neural sequence labelling model to the task of de-identification of medical records. 
One appealing property of those works is that they can achieve excellent performance with a unified architecture and without task-specific feature engineering. It remains unclear that whether such works can be used for tasks without large amounts of training data. Several works such as Yang et al.~\shortcite{YangZ17} and Lee et al.~\shortcite{LeeJY18} proposed applying transfer learning to NER using a limited number of training corpora. Nevertheless, a large dataset that has same labels as the small training dataset is required for transfer learning, which is hard to obtain in the field of cybersecurity. In this paper, we introduce several spelling features which use no expert knowledge of cybersecurity to the neural model, and achieve an excellent performance even using a small dataset for training.

\section{Conclusions}
\label{sec:conc}

 To conclude, in this paper, we propose a neural based sequence labelling model capable of identifying IOCs automatically from APT security reports. Utilizing an attention mechanism and several spelling features, we find that the proposed model can correctly identify low frequency IOCs with a small amount of training corpora. Based on the evaluation results of our experiments on English APT reports, our proposed approach performs better than other sequence labelling models with an average precision of 90.4\% and recall of 87.2\%.
 
 To avoid the problem caused by the spelling features described in Section~\ref{sec:result}, one of our significant future work is to integrate several context features. 
 Another important future work is to adapt the proposed model to another new language. 
 Even though security articles are written in different languages, most of the IOCs are written in English. Our preliminary experiments demonstrate that models trained with English texts can identify parts of IOCs from a Chinese text using cross-lingual words embedding obtained by the work of Duong el al~\shortcite{Duong16}. It can be a quick way to adapt the model to a new language with minimal or no data, and the performance of the proposed model is expected to be improved by extending the training dataset using multilingual corpora.

%

\bibliographystyle{acl}
\bibliography{long}

\end{document}